# Modeling And Simulation Of Prolate Dual-Spin Satellite Dynamics In An Inclined Elliptical Orbit: Case Study Of Palapa B2R Satellite

J. Muliadi[*], S.D. Jenie[†] and A. Budiyono[‡]

*Abstract— In response to the interest to re-use Palapa B2R satellite nearing its End of Life (EOL) time, an idea to incline the satellite orbit in order to cover a new region has emerged in the recent years. As a prolate dual-spin vehicle, Palapa B2R has to be stabilized against its internal energy dissipation effect. This work is focused on analyzing the dynamics of the reusable satellite in its inclined orbit. The study discusses in particular the stability of the prolate dual-spin satellite under the effect of perturbed field of gravitation due to the inclination of its elliptical orbit. Palapa B2R physical data was substituted into the dual-spin's equation of motion. The coefficient of zonal harmonics $J_2$ was induced into the gravity-gradient moment term that affects the satellite attitude. The satellite's motion and attitude were then simulated in the perturbed gravitational field by $J_2$, with the variation of orbit's eccentricity and inclination. The analysis of the satellite dynamics and its stability was conducted for designing a control system for the vehicle in its new inclined orbit.*

*Abstrak— Sebagai reaksi atas adanya minat dari beberapa pihak untuk mengoperasikan kembali satelit Palapa B2R yang mendekati masa akhir operasinya (End of Life), suatu ide untuk menginklinasi orbit satelit telah mengemuka pada beberapa tahun terakhir. Sebagai sebuah wahana* dual-spin prolate, *Palapa B2R harus distabilkan terhadap efek disipasi energi internal. Paper ini berkonsentrasi pada analisis dinamik dari satelit pada orbit barunya yang terinklinasi. Studi yang dilakukan mendiskusikan secara khusus kestabilan dari satelit dual-spin prolate dalam pengaruh efek dari medan gravitasi terganggu akibat inklinasi dari orbit eliptiknya. Data fisik Palapa B2R disubstitusikan ke dalam persamaan gerak dual-spin. Koefisien harmonik zonal $J_2$ diinduksikan ke persamaan momen gradient gravitasi yang mempengaruhi sikap satelit. Gerak dan sikap satelit kemudian disimulasikan dengan variasi eksentrisitas dan inklinasi orbitnya. Analisis dinamika dan kestabilan satelit dilakukan untuk keperluan perancangan sistem kendali wahana pada orbit barunya yang terinklinasi.*
*Kata kunci—dual-spin, dual-spin prolate, dinamika satelit, simulasi*

[*] Department of Aeronautics and Astronautics, ITB, mie_pn@yahoo.com
[†] Department of Aeronautics and Astronautics, ITB, sdjenie@ae.itb.ac.id
[‡] Formerly with Department of Aeronautics and Astronautics, ITB, presently with Department of Aerospace-IT Fusion Engineering, Konkuk University, Korea, the author to whom all correspondence to be addressed budiyono@alum.mit.edu

## 1. INTRODUCTION

A semi-rigid body is stable only when spinning about its major axis. In a related study, Bracewell and Garriott (Ref. [8] pp. 62-64) concluded that the four turnstile wire antennae of Explorer I were dissipating energy; thus, causing a transfer of body spin axis from the minimum inertia (*prolate*) to a transverse axis of maximum inertia (*oblate*). To meet this stability criterion, most of early dual-spin vehicles were designed in an oblate configuration.

As a case of study, this work used Palapa B2R physical data to analyze the dynamics of the vehicle. Palapa B2R is a communication satellite of Indonesia. In its orbit, it was operated by *PT Telkom* (Indonesian State Telecommunication Company). Near the satellite's End of Life (EOL) time, several Africans and Polynesians countries have shown interest to buy and re-use Palapa B2R. Because of those countries' location in the southern latitudes, an idea emerged to incline the satellite's orbit. The current paper elaborates the analysis of the vehicle dynamics in its inclined orbit.

## 2. REFERENCE COORDINATE SYSTEM

### 2.1. Body Reference Coordinate System (Body Axes)

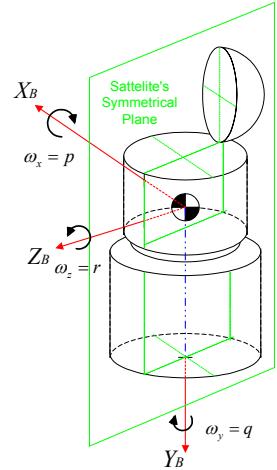

Fig. 1 Platform Axis Components

The definition of Platform and Body axes is well-defined in the literature. Fig. 1 illustrated those axes with their origin at the satellite's c.g. while Fig. 2 showed the axes in the space. In this paper, the Platform Axis Components will be identified as Body Reference Coordinate System or Body axis.

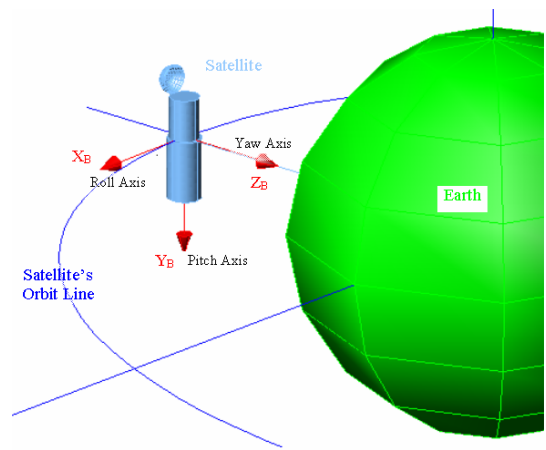

Fig. 2 Body Reference Coordinate System

## 2.2. Stability Reference Coordinate System (Stability Axes)

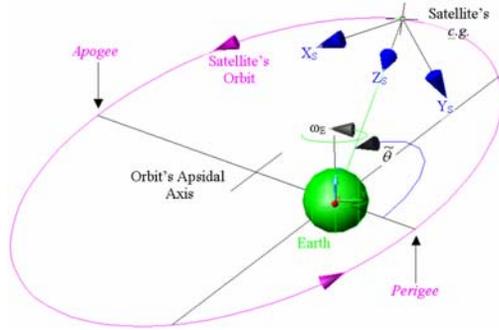

Fig. 3 Stability Reference Coordinate System

Stability Reference Coordinate System (Stability Axes) was defined as a set of local horizon axes for the satellite. It is a target axes for the satellite's Body Axes to point its antennae to the Earth.. All these axes are presented in Fig. 3.

## 2.3. Inertial Reference Coordinate System (Inertial Axes)

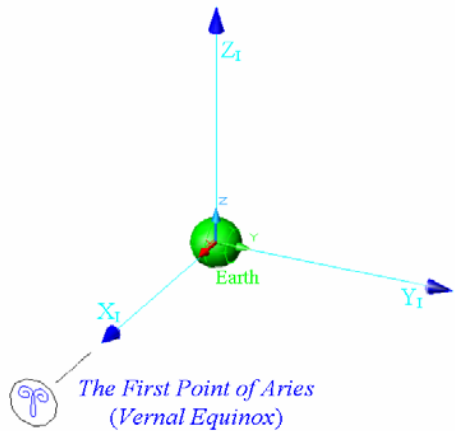

Fig. 4 Inertial Reference Coordinate System

Inertial Reference Coordinate System (Inertial Axes) is defined as a *geocentric non-rotating equatorial reference frame* with $Z_I$ axis which coincides with the rotation axis of the Earth and points to the North Pole; the $X_I$ axis lies in equatorial plane and points towards the *vernal equinox*. The $Y_I$ axis completes a right-handed Cartesian frame of reference. In this Inertial Axis, Newton's laws of motion are valid for the satellite's translation and rotation.

## 3. Euler Angles (Orientation Angles)

### 3.1. Orientation of Body Axes in Inertial Axes

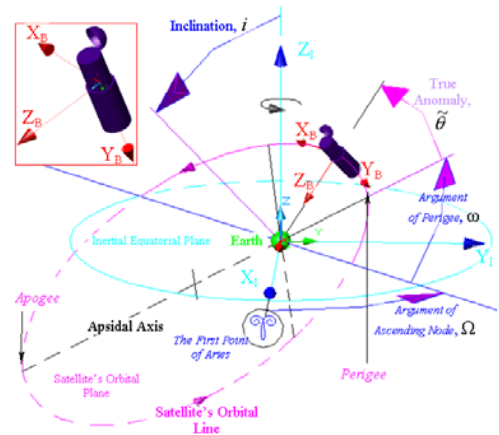

Fig. 5 Orientation of Body Axes in Inertial Axes

In order to describe the attitude of the satellite with respect to the Inertial Axes (Fig. 5), the Euler angles are used.

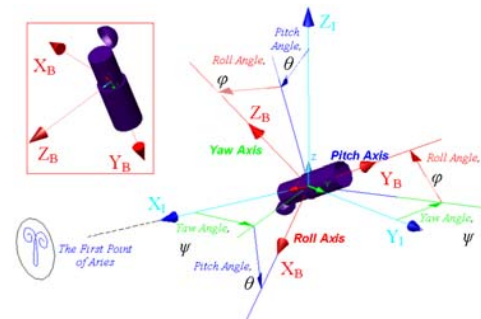

Fig. 6 Euler Angles of Body Axis in Inertial Axis

The yaw angle $\psi$, pitch angle $\theta$ and roll angle $\varphi$, respectively defines the angle of rotation in Z-, Y- and X- axis of the Body frame with respect to its nominal condition in Inertial Axes.

These angles are shown in Fig. 6.

### 3.2. Orientation of Body Axes in Stability Axes

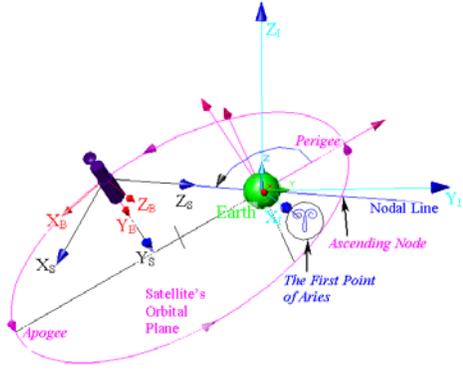

Fig. 7 Deviation of Body Axes from Stability Axes

Euler Angles of Body Axes with respect to Stability Axes are defined to describe the attitude perturbation from its local horizon (its stationary or nominal condition). If the satellite deviates to large, the antennae will point away from the Earth. The Euler Angles are:

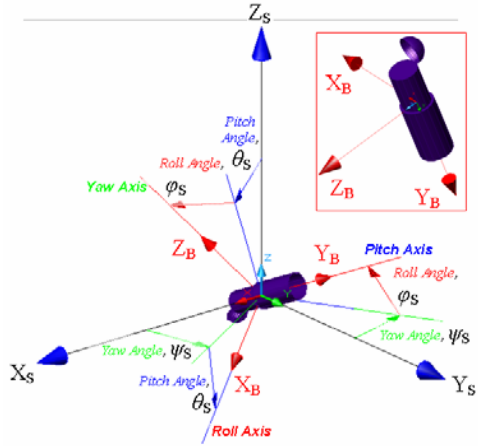

Fig. 8 Euler Angles of Body Axis with Respect to Stability Axis

*Yaw, Pitch and Roll Angle Deviation* ($\psi_S$, $\theta_S$, and $\varphi_S$), which respectively denotes the angle of perturbation because of the rotation in Z-, Y-, and X-axis of the Body frame with respect to Stability Axes. These angles are shown in Fig. 8.

### 4. GRAVITY GRADIENT MOMENT

Agrawal derived an expression for Gravity Gradient Moment in Axially Symmetric Spacecraft at Equatorial Circular Orbit (Ref [1] pp. 131-133) with this equation,

$$\mathbf{M}_G = 3 \cdot \frac{\mu_E}{R_0^3} \begin{Bmatrix} (I_Z^B - I_Y^B) \cdot \phi_S \\ (I_Z^B - I_X^B) \cdot \theta_S \\ 0 \end{Bmatrix}$$

Eq. 1

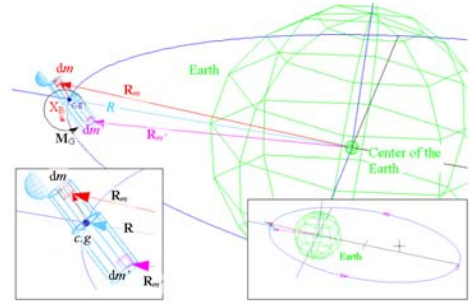

Fig. 9 Concept of Gravity Gradient Moment

It this paper the spacecraft will be treated as a non-axially symmetric vehicle. In addition, the satellite will be operated in inclined elliptical orbit, which means non-equatorial and non-circular orbit. So, by following and combining Kaplan's (Ref. [8] pp. 199-204) and Agrawal's techniques in deriving the equation of Gravity Gradient Moment for Spacecraft, the authors derived the equation of Gravity Gradient Moment for Prolate Dual-Spin Satellite in its inclined elliptical orbit.

The gravitational force (d$\mathbf{F}_G$) corresponding to a differential element of mass, d*m*, shown in **Fig. 9**, is

$$d\mathbf{F}_G = \mathbf{g} \cdot dm$$

Eq. 2

where **g** denotes gravity vector at d*m*. The unsymmetrical mass distribution of the Earth induced a *zonal harmonic coefficient* ($J_k$, Ref [8] pp. 273-282) that perturbs the homogenous of the Earth gravity's field. In this work, the oblateness of the Earth induced the *zonal harmonic coefficient* that is limited to second order, $J_2$. The gravity vector equation at a point in space is

$$\mathbf{g} = \frac{-\mu_E}{R^3} \cdot \left[ 1 - \frac{3}{2} \cdot J_2 \cdot \left[ \frac{Re}{R} \right]^2 \cdot \left( 5 \cdot \frac{R_{Z_E}^2}{R^2} - 1 \right) \right] \cdot \mathbf{R}$$

Eq. 3

With $\mu_E$ = Earth gravitational parameter; $R$ and $\mathbf{R}$ = the distance from the center of the Earth in scalar and vector notation; $R_{Z_E}$ = the height measured perpendicular from the Earth equatorial plane; Re = the radius of the Earth equator.

Continuing to the moment equation,

$$\mathbf{M}_G = \int_B \mathbf{r} \times d\mathbf{F}_G = \int_B \mathbf{r} \times \mathbf{g} \cdot dm$$

Eq. 4

while the position of differential element of mass, d*m*, in **Fig. 9** is

$$\mathbf{R}_m = \mathbf{R} + \mathbf{r}$$

Eq. 5

Therefore, substituting Eq. 5 to replace $\mathbf{R}$ in Eq. 3, then inserting the result to Eq. 4, tranforms Eq. 4 into:

$$\mathbf{M}_G = -\int_B \mathbf{r} \times$$
$$\left\{ \frac{\mu_E}{|\mathbf{R}+\mathbf{r}|^3} \cdot \left[ 1 - \frac{3}{2} \cdot J_2 \cdot \frac{Re^2}{|\mathbf{R}+\mathbf{r}|^2} \cdot \left( 5 \cdot \frac{R_{Z_E}^2}{|\mathbf{R}+\mathbf{r}|^2} - 1 \right) \right] \cdot (\mathbf{R}+\mathbf{r}) \right\} \cdot dm$$

Eq. 6

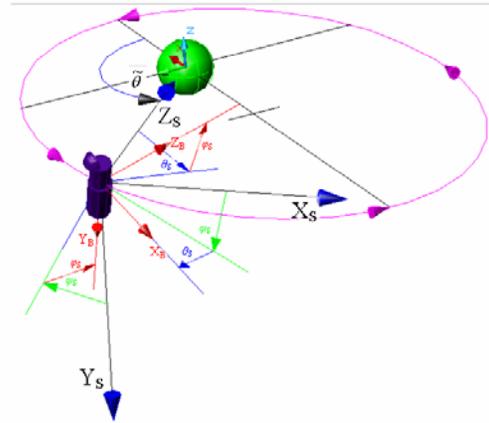

Fig. 10 Deviation from Steady Attitude

The use of Agrawal's technique to work with Euler Angles between Body Axis and Stability Axis see also **Fig. 8**), yields (eq. 3.101, Ref. [1] p. 132),

$$R_X = R \cdot \sin \theta_S$$
$$R_Y = -R \cdot (\sin \phi_S \cdot \cos \theta_S)$$
$$R_Z = -R \cdot (\cos \phi_S \cdot \cos \theta_S)$$

Eq. 7

Binomial Expansion is applied into Eq. 6 and then Eq. 7 was inserted into the result to give Gravity Gradient Moment Equation. By linearizing the equations, the Linearized Gravity Moment Equations read,

$$M_{GX} = G_X \cdot \phi_S$$
$$M_{GY} = G_Y \cdot \theta_S$$
$$M_{GZ} = G_Z \cdot \theta_S$$

Eq. 8

Where coefficients of $\phi_S$ and $\theta_S$ are:

$$G_X = g_\mu \bullet (I_Z^B - I_Y^B)$$
$$G_Y = g_\mu \bullet (I_Z^B - I_X^B)$$
$$G_Z = g_\mu \bullet I_{YZ}^B$$

Eq. 9

And coefficient $g_\mu$ is,

$$g_\mu = \left\{ 3 \cdot \frac{\mu_E}{R^3} + \left( \frac{105}{2} \cdot \frac{\mu_E}{R^7} \cdot J_2 \cdot Re^2 \cdot R_{Z_1}^2 \right) \right.$$
$$\left. - \left( \frac{15}{2} \cdot \frac{\mu_E}{R^5} \cdot J_2 \cdot Re^2 \right) \right\}$$

The $R_{Z_E}$ variable is the component of satellite's position at Z-Axis in Earth. The factor of inclination was induced in $R_{Z_I}$ variable, because the height of the satellite will be varying in the inclined orbit. The factor of eccentricity was induced in R variable. For $e>0$, the value of $R$ is varying along the orbit.

## 5. ORBITAL MOTIONS

### 5.1. Parameters of Keplerian Orbit

Astronomy defines 6 quantities to describe the orbit and position of heavenly body, namely a, e, i, $\Omega$, $\omega$, and $\tau$. The definition of those parameters can be found in many textbook in orbital mechanics. Fig. 12 describes the geometry of the orbital parameters.

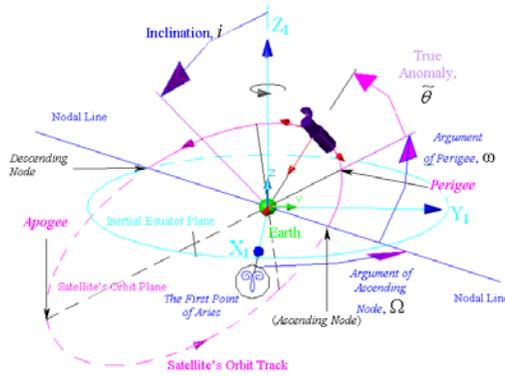

Fig. 11 Orbital Plane Orientation with respect to Inertial Axes

### 5.2. Orbital Motion Equations

Simulation of the orbital motion uses scalar differentials equation in $X_I$-axis, $Y_I$-axis and $Z_I$-axis as follows:

$$\ddot{R}_{XI} = \frac{\mu_E}{R^3} \cdot R_{XI}$$

$$\ddot{R}_{YI} = \frac{\mu_E}{R^3} \cdot R_{YI}$$

$$\ddot{R}_{ZI} = \frac{\mu_E}{R^3} \cdot R_{ZI}$$

**Eq. 10**

**Eq. 11**

The equation was derived by relating Newton's Second Law of Motion and Newton's Law for Gravitation. To sketch the satellite's orbit, Eq. 11 were integrated two times with 6 initial values, initial velocity $\dot{R}_{XI}$, $\dot{R}_{YI}$, $\dot{R}_{ZI}$, and 3 initial position $R_{XI}$, $R_{YI}$, and $R_{ZI}$, when t=0. From Jenie (Ref. [6]), initial velocity can be expressed by

$$\begin{Bmatrix} V_{XI} \\ V_{YI} \\ V_{ZI} \end{Bmatrix} = \mathbf{C}_I^{Orb} \begin{bmatrix} \cos\tilde{\theta} & \sin\tilde{\theta} & 0 \\ -\sin\tilde{\theta} & \cos\tilde{\theta} & 0 \\ 0 & 0 & 1 \end{bmatrix} \begin{Bmatrix} \sqrt{\frac{\mu_E}{a(1-e^2)}} \bullet e\sin\tilde{\theta} \\ \sqrt{\frac{\mu_E}{a(1-e^2)}} \bullet (1+e\cos\tilde{\theta}) \\ 0 \end{Bmatrix}$$

**Eq. 12**

and initial position can be expressed by

$$\begin{Bmatrix} R_{XI} \\ R_{YI} \\ R_{ZI} \end{Bmatrix} = \mathbf{C}_I^{Orb} \begin{Bmatrix} \left(\frac{a(1-e^2)}{1+e\cos\tilde{\theta}}\right) \bullet \cos\tilde{\theta} \\ \left(\frac{a(1-e^2)}{1+e\cos\tilde{\theta}}\right) \bullet \sin\tilde{\theta} \\ 0 \end{Bmatrix}$$

**Eq. 13**

with transformation matrix $\mathbf{C}_I^{Orb}$, and its elements as follows:

$$\mathbf{C}_I^{Orb} = \begin{bmatrix} C_{11} & C_{12} & C_{13} \\ C_{21} & C_{22} & C_{23} \\ C_{31} & C_{32} & C_{33} \end{bmatrix}$$

$C_{11} = (\cos\omega\cos\Omega - \sin\omega\cos i\sin\Omega)$
$C_{12} = (-\sin\omega\cos\Omega - \cos\omega\cos i\sin\Omega)$
$C_{13} = (\sin\Omega\sin i)$

$C_{21} = (\cos\omega\sin\Omega + \sin\omega\cos i\cos\Omega)$
$C_{22} = (-\sin\omega\sin\Omega + \cos\Omega\cos\omega\cos i)$
$C_{23} = (-\sin i\cos\Omega)$

$C_{31} = (\sin\omega\sin i)$
$C_{32} = (\cos\omega\sin i)$
$C_{33} = (\cos i)$

When simulating the satellite's orbit, the orbital parameters are
$\tilde{\theta}_0 = 0^0$, $\Omega_0 = 0^o$, $\omega_0 = 0^o$

$a_0 = 8078.14$ km
for circular orbit
$R = 1700$ km above the sea level;
for elliptic orbit with e=0.2, perigee,
$R_{\text{perigee}} = 85$ km above the sea level.

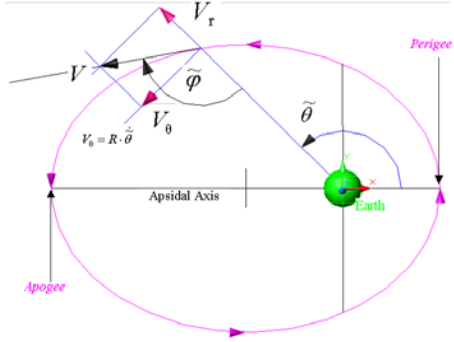

Fig. 102 Velocities in Orbit

In the state space model for attitude dynamics, the satellite transversal velocity that perpendicular to its radius to the Earth, $V_\theta$, will be needed.

A. E. Roy (Ref. [9] p. 81) relates $\sin\widetilde{\varphi}$ in **Fig. 10** with eccentricity and semimajor axis as follows:

$$\sin\widetilde{\varphi} = \left[\frac{a^2 \cdot (1-e^2)}{R \cdot (2 \cdot a - R)}\right]^{\frac{1}{2}}$$

**Eq. 14**

With Eq. 14, the component of velocity in theta direction can be stated as

$$V_\theta = V \cdot \sqrt{\frac{a^2 \cdot (1-e^2)}{R \cdot (2 \cdot a - R)}}$$

**Eq. 15**

## 6. DYNAMICS OF PROLATE DUAL SPIN SATELLITE

Palapa B2R is a *prolate* configuration Communication Satellite (HS 376). In order to stabilize its attitude and pointing direction, Palapa B2R uses its rotor spinning. Control moments were produced by the angular acceleration and deceleration of the rotor's spin. Because of the rotor's spinning motion and the configuration of satellites inertia, the satellite's motion in the yaw mode is coupled with its roll mode. In addition, by the imbalance from the satellite's antennae reflector configuration ($I_{YZ}$), the satellite's motion in the pitch mode is coupled with its yaw mode.

### 6.1. Dynamic Equations of Motion

Bryson has shown the equation of spacecraft motion in De-Spin Active Nutation Damping in Ref. [2 pp. 62-68]. In this work, the author will insert gravity gradient moment as an external moment that perturbs the satellite's attitude. Using Bryson's technique in deriving the De-Spin Active Nutation Damping equations, the author adds several modifications. The results are:

**The Dynamics Equation at X-axis**,
$$(I_X + I_T) \cdot \dot{p} - (I_S \cdot \Omega_{R0}) \cdot r = M_X$$

**Eq. 16**

**The Dynamics Equation at Y-axis**,
$$I_Y \cdot \dot{q} + I_{YZ} \cdot \dot{r} + \left(\frac{N \cdot K_V}{R_{d.c.}} + c\right) \cdot \left(\frac{I_Y}{I_S} + 1\right) \cdot q$$
$$+ \left(\frac{N \cdot K_V}{R_{d.c.}} + c\right) \cdot \frac{I_{YZ}}{I_S} r + \frac{N}{R_{d.c.}} \cdot \delta e = M_Y$$

**Eq. 17**

**The Dynamics Equation at Z-axis**,
$$I_{YZ} \cdot \dot{q} + (I_Z + I_T) \cdot \dot{r} + (I_S \cdot \Omega_{R0}) \cdot p = M_Z$$

**Eq. 18**

### 6.2. Dynamic Equations with Gravity Gradient Moments

Substituting $M_X$ in Eq. 16 with $M_{G_X}$ in Eq. 8 yields the Dynamics Equation at X-axis in Inclined Elliptical Orbit,

$$(I_X + I_T) \cdot \dot{p} - (I_S \cdot \Omega_{R0}) \cdot r$$
$$- G_X \cdot \phi_S = 0$$

**Eq. 19**

Substituting $M_Y$ in Eq. 17 with $M_{G_Y}$ in Eq. 8 yields the Dynamics Equation with

Gravity Gradient Moments at Y-axis in Inclined Elliptical Orbit,

$$I_Y \cdot \dot{q} + I_{YZ} \cdot \dot{r} + \left(\frac{N \cdot K_V}{R_{d.c.}} + c\right) \cdot \left(\frac{I_Y}{I_S} + 1\right) \cdot q$$
$$+ \left(\frac{N \cdot K_V}{R_{d.c.}} + c\right) \cdot \frac{I_{YZ}}{I_S} r - G_Y \cdot \theta_S + \frac{N}{R_{d.c.}} \cdot \delta e = 0$$

**Eq. 20**

Finally, substituting $M_Z$ in Eq. 18 with $M_{G_Z}$ in Eq. 8 yields the Dynamics Equation at Z-axis in Inclined Elliptical Orbit,

$$(I_X + I_T) \cdot \dot{p} - (I_S \cdot \Omega_{R0}) \cdot r$$
$$- G_X \cdot \phi_S = 0$$

**Eq. 21**

## 6.3. Kinematics Equations of Motion

### 6.3.1. Kinematics Equations in Inertial Axes

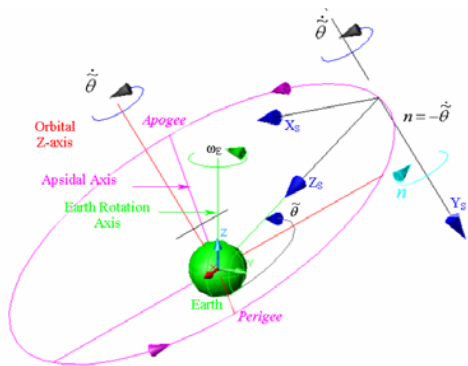

Fig. 113 Rotation of Local Horizon Axes (Stability Axes), *n*

Bryson has shown that the equation of kinematics in relationship between Euler Angle Rates and Angular Velocity (Ref. [2] pp. 8-9) can be approximated as

$$\dot{\phi} \cong p + n \cdot \psi$$
$$\dot{\theta} \cong q + n$$
$$\dot{\psi} \cong r - n \cdot \phi$$

**Eq. 22**

where *n* = orbital angular velocity. These kinematics equations are derived in Inertial Axis by Newton's first and second law of motion.

### 6.3.2. Kinematics Equations in Inertial Axes

In the beginning of simulation, the satellite's antennae still points to the Earth for a moment. By the inertia (the Newton first law of motion) the satellite's attitude will drift at rotor spin axis with drift rate equals to *orbital angular rate*, $\dot{\tilde{\theta}}$, or equals to initial value of *n*, $n_0$. Therefore, to measure the deviation in Stability Axes, the kinematics equation needs to be reduced by the effect of Local Horizon Axes (Stability Axes) initial angular drift with respect to Inertial Axes, $n_0$. Following Byson's techniques in deriving the kinematics equation, the *Kinematics Equation of Dual Spin Satellite in Stability Axis* can be written as follows

$$\dot{\phi}_S \cong p + \delta n \cdot \psi_S$$
$$\dot{\theta}_S \cong q + \delta n$$
$$\dot{\psi}_S \cong r - \delta n \cdot \phi_S$$

**Eq. 23**

where $\delta n = (n - n_0)$.

## 6.4. State Space Model of Prolate Dual-Spin

Combining The Dynamics Equation in Inclined Elliptical Orbit and Kinematics Equation of Dual Spin Satellite in Stability Axis yields State Space Model for Dual-Spin Satellite in Stability Axes as follows,

$$\begin{Bmatrix} \dot{p} \\ \dot{q} \\ \dot{r} \\ \dot{\phi}_S \\ \dot{\theta}_S \\ \dot{\psi}_S \end{Bmatrix} = [\mathbf{A}] \bullet \begin{Bmatrix} p \\ q \\ r \\ \phi_S \\ \theta_S \\ \psi_S \end{Bmatrix} + [\mathbf{B}] \bullet \begin{Bmatrix} \delta e \\ \delta n \end{Bmatrix}$$

**Eq. 24**

where the [**A**] and [**B**] matrices are:

$$[A] = \begin{bmatrix} 0 & 0 & A_{13} & A_{14} & 0 & 0 \\ A_{21} & A_{22} & A_{23} & 0 & A_{25} & 0 \\ A_{31} & A_{32} & A_{33} & 0 & A_{35} & 0 \\ 1 & 0 & 0 & 0 & 0 & \delta n \\ 0 & 1 & 0 & 0 & 0 & 0 \\ 0 & 0 & 1 & -\delta n & 0 & 0 \end{bmatrix}$$

$$[B] = \begin{bmatrix} 0 & 0 \\ B_{21} & 0 \\ B_{31} & 0 \\ 0 & 0 \\ 0 & 1 \\ 0 & 0 \end{bmatrix}$$

By defining $\Delta_I$ as follows,
$$\Delta_I = (I_Y I_Z + I_Y I_T - I_{YZ}^2)$$

the [A] matrix elements in 1$^{st}$ line are:
$$A_{13} = \frac{-1}{(I_X + I_T)} \cdot (I_S \cdot \Omega_{R0})$$
$$A_{14} = \frac{-1}{(I_X + I_T)} \cdot G_X$$

the [A] matrix elements in 2$^{nd}$ line are:
$$A_{21} = \frac{-I_{YZ}}{\Delta_I} \cdot (I_S \cdot \Omega_{R0})$$
$$A_{22} = \frac{I_Z + I_T}{\Delta_I} \cdot \left(\frac{N \cdot K_V}{R_{d.c.}} + c\right) \cdot \left(\frac{I_Y}{I_S} + 1\right)$$
$$A_{23} = \frac{I_Z + I_T}{\Delta_I} \cdot \left(\frac{N \cdot K_V}{R_{d.c.}} + c\right) \cdot \left(\frac{I_{YZ}}{I_S}\right)$$
$$A_{25} = \frac{-(I_Z + I_T)}{\Delta_I} \cdot G_Y + \frac{I_{YZ}}{\Delta_I} \cdot G_Z$$

the [A] matrix elements in 3$^{rd}$ line are:
$$A_{31} = \frac{I_Y}{\Delta_I} \cdot (I_S \cdot \Omega_{R0})$$
$$A_{32} = \frac{-I_{YZ}}{\Delta_I} \cdot \left(\frac{N \cdot K_V}{R_{d.c.}} + c\right) \cdot \left(\frac{I_Y}{I_S} + 1\right)$$
$$A_{33} = \frac{-I_{YZ}^2}{\Delta_I} \cdot \left(\frac{N \cdot K_V}{R_{d.c.}} + c\right) \cdot \left(\frac{1}{I_S}\right)$$
$$A_{35} = \frac{I_{YZ}}{\Delta_I} \cdot G_Y - \frac{-I_Y}{\Delta_I} \cdot G_Z$$

the [B] matrix elements in 1$^{st}$ column are:
$$B_{21} = \frac{I_Z + I_T}{\Delta_I} \cdot \left(\frac{N}{R_{d.c.}}\right)$$

$$B_{31} = \frac{-I_{YZ}}{\Delta_I} \cdot \left(\frac{N}{R_{d.c.}}\right)$$

and

$$n = -\dot{\tilde{\theta}} = \frac{-V_\theta}{R}$$
$$\delta n = (n - n_0)$$
$$= \dot{\tilde{\theta}}_0 - \dot{\tilde{\theta}} = \frac{(V_\theta)_0}{R_0} - \frac{(V_\theta)}{R}$$

## 7. RESULTS OF OPEN LOOP SIMULATION AND INTERPRETATIONS

The values of [A]'s and [B]'s elements are shown in Eq. 25. The value of $A_{14}$, $A_{25}$, $A_{35}$ and $\delta n$ will be time-varying for elliptical or inclined orbit. However, the elements of [A] are constant value only for circular orbit at equatorial plane.

$$A = \begin{bmatrix} 0 & 0 & 3.7113 & A_{14} & 0 & 0 \\ 0.49773 & -9.7138 \times 10^{-4} & -3.4402 \times 10^{-5} & 0 & A_{25} & 0 \\ -4.0326 & 3.3636 \times 10^{-5} & -1.1912 \times 10^{-6} & 0 & A_{35} & 0 \\ 1 & 0 & 0 & 0 & 0 & \delta n \\ 0 & 1 & 0 & 0 & 0 & 0 \\ 0 & 0 & 1 & -\delta n & 0 & 0 \end{bmatrix}$$

$$B = \begin{bmatrix} 0 & 0 \\ -5.1218 \times 10^{-4} & 0 \\ 1.7735 \times 10^{-5} & 0 \\ 0 & 0 \\ 0 & 1 \\ 0 & 0 \end{bmatrix}$$

Eq. 25

### 7.1. Simulation in Longitudinal Mode

#### 7.1.1. Effect of Eccentricity in Inclined Orbit (*i* = 30°)

◆ Plot of $\theta_S$ because of impulsive input $\delta e_{ref}$

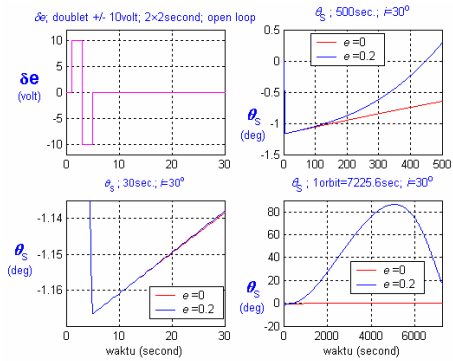

Fig. 124 Plot of $\theta_S$; input $\delta e_{\text{ref}}$ ; $i$=30deg

### ◆ Plot of $\theta_S$ due to elliptic orbital drift input $\delta n$

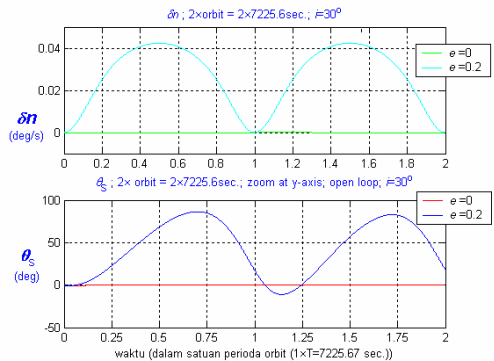

Fig. 135 Plot of $\theta_S$ and $\delta n$; $e$=0 & $e$=0.2; 2 Orbital Periode; $i$=30deg

After impulsive disturbance were applied, Gravity Gradient Moment induced subsidence mode in circular orbit. While the increment of eccentricity, drove that subsidence mode into long periodic oscillation mode, which had the same period with orbital period. This long periodic oscillation mode oscillates from –20º to +90º as superposition of many oscillation mode.

#### 7.1.2. Effect of Inclination in Elliptic Orbit ($e$ = 0.2)

◆ Plot of $\theta_S$ due to impulsive input $\delta e_{\text{ref}}$

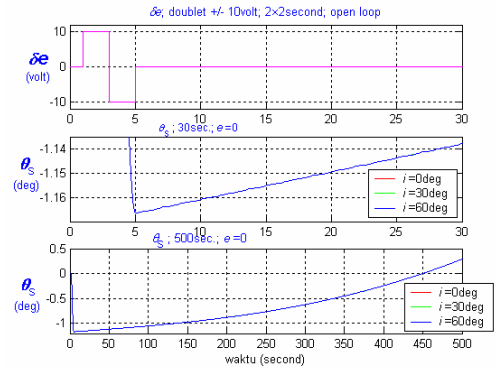

**Fig**. 146 Plot of $\theta_S$ $i$=0deg, 30deg dan 60deg; $e$=0.2

### ◆ Plot of $\theta_S$ due to elliptic orbital drift input $\delta n$

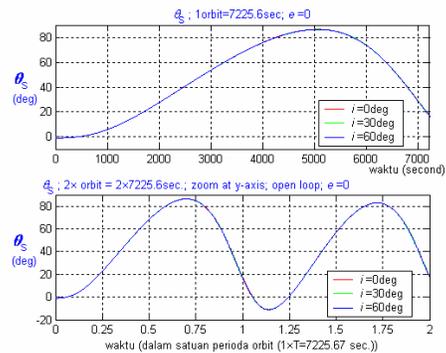

Fig. 157 Plot of $\theta_S$; $i$=0deg, 30deg dan 60deg;1Orbit and 2 Orbits; $e$=0.2

Inclination increment effect in longitudinal motion mode was negligible for Dual-Spin Satellite. For inclination at 0°, 30°, and 60°, the graphic curves for $\theta_S$ were almost aligned.

#### 7.2. Simulation in Lateral Mode

##### 7.2.1. Effect of Eccentricity in Inclined Orbit ($i$ = 30°)

◆ Plot of $\varphi_S$ due to impulsive input $\delta e_{\text{ref}}$

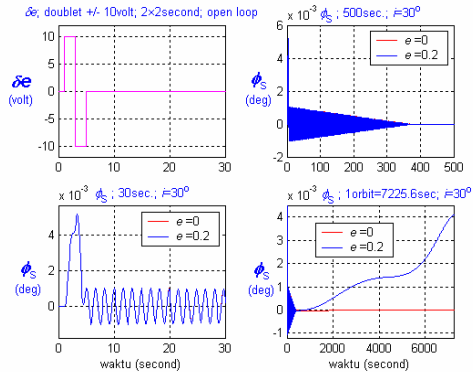

Fig. 18 Plot of $\varphi_S$; $e$=0 and $e$=0.2; $i$=30deg

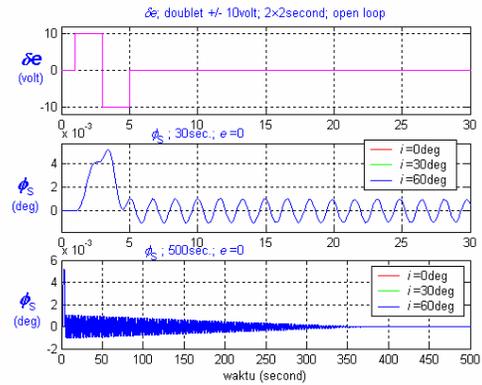

Fig. 170 Plot of $\varphi_S$; $i$=0deg, 30deg dan 60deg; $e$=0.2

◆ **Plot of $\varphi_S$ due to elliptic orbital drift input $\delta n$**

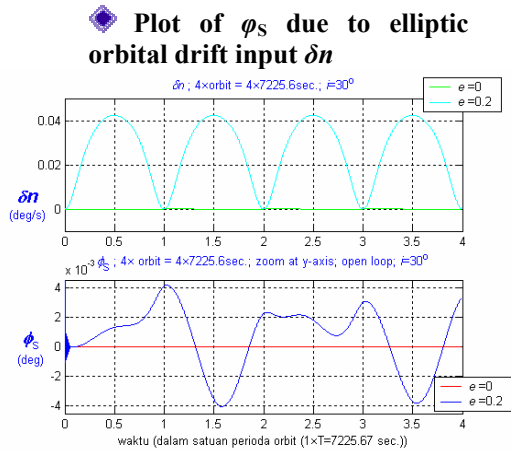

Fig. 16 Plot of $\varphi_S$ and $\delta n$; $e$=0 & $e$=0.2; 4 Orbits; $i$=30deg

◆ **Plot of $\varphi_S$ due to elliptic orbital drift input $\delta n$**

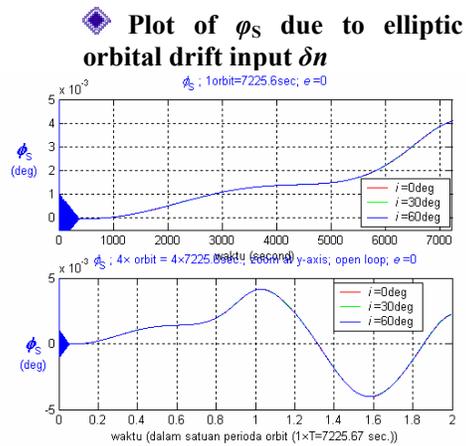

Fig. 18 Plot of $\varphi_S$ and $\delta n$; $i$=0deg, 30deg and 60deg; 2 Orbits $e$=0.2

After impulsive disturbance were applied, Gravity Gradient Moment induced *roll-librations* mode in for $\varphi_S$, which oscillates from $-1\times10^{-3}$ to $+1\times10^{-3}$ deg. In circular orbit, this *roll-librations* mode in for $\varphi_S$ was damped after ±400 sec. While in elliptic orbit, the eccentricity induced long periodic oscillation mode, which oscillates from $-4\times10^{-3}$ to $+4\times10^{-3}$ deg.

### 7.2.2. Effect of Inclination in Elliptic Orbit (*e* = 0.2)

◆ **Plot of $\varphi_S$ due to impulsive input $\delta e_{ref}$**

Inclination increment effect in longitudinal motion mode was negligible for Dual-Spin Satellite. For inclination at 0°, 30°, and 60°, the graphic curves for $\varphi_S$ were almost aligned.

### 7.3. Simulation in Directional Mode

#### 7.3.1. Effect of Eccentricity in Inclined Orbit (*i* = 30°)

◆ **Plot of $\psi_S$ due to impulsive input $\delta e_{ref}$**

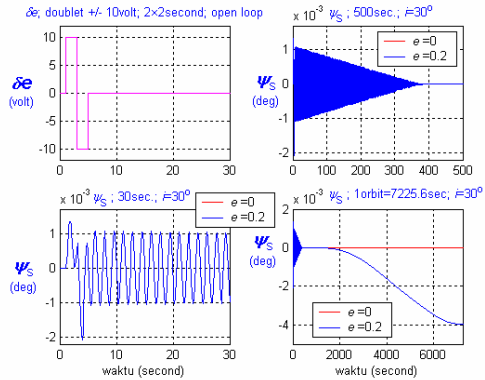

Fig. 19 Plot of $\psi_S$; $e$=0 and $e$=0.2; $i$=30deg

### ◆ Plot of $\psi_S$ due to elliptic orbital drift input $\delta n$

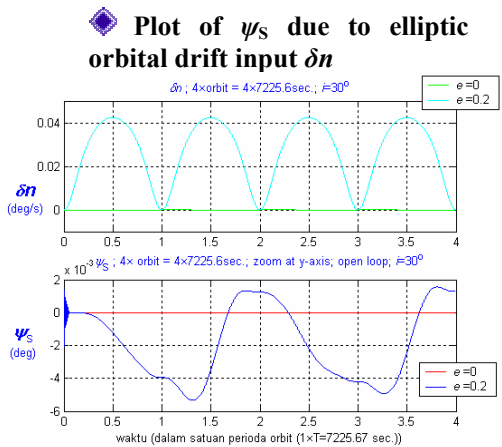

Fig. 20 Plot of $\psi_S$ and $\delta n$; $e$=0 & $e$=0.2; 4 Orbits $i$=30deg

After impulsive disturbance were applied, Gravity Gradient Moment induced *yaw-librations* mode in for $\varphi_S$. In circular orbit, this *yaw-librations* mode in for $\psi_S$ was damped after ±400 sec. In the elliptic orbit, the eccentricity induces long periodic oscillation mode, which oscillates from $-5.5 \times 10^{-3}$ to $+1.5 \times 10^{-3}$ (deg).

#### 7.3.2. Effect of Inclination in Elliptic Orbit (e = 0.2)

### ◆ Plot of $\psi_S$ due to impulsive input $\delta e_{ref}$

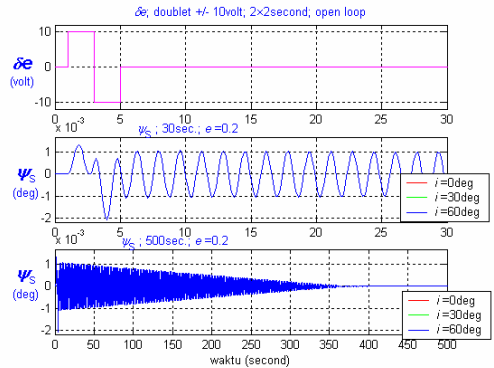

Fig. 21 Plot of $\psi_S$; $i$=0deg, 30deg and 60deg; $e$=0.2

### ◆ Plot of $\psi_S$ due to elliptic orbital drift input $\delta n$

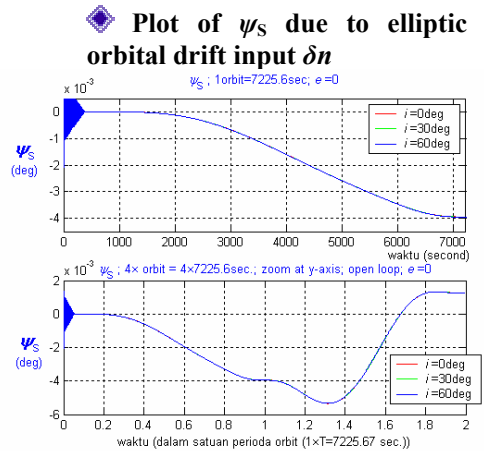

Fig. 22 Plot of $\psi_S$; $i$=0deg, 30deg and 60deg;1 Orbit and 2 Orbits; $e$=0.2

Inclination increment effect in longitudinal motion mode was negligible for Dual-Spin Satellite. For inclination at 0°, 30°, and 60°, the graphic curves for $\psi_S$ were almost aligned.

## 8. CONCLUDING REMARKS

In Open-Loop simulation, the librations and subsidence mode are present in longitudinal, lateral and directional motion. However, in lateral and directional motion, the Gravity Gradient moment induced a divergence mode if the simulation were run for more than 4 orbital periods (1 Orbital Periods is 7225 sec.).

The effect of Gravity Gradient Moments is destabilizing the lateral and directional motion for the dual-spin satellite. In reverse, the effect of Gravity

Gradient Moments is stabilizing longitudinal motion. The effect of increasing the orbital eccentricity, *e*, is the presence of the long period oscillation mode in the longitudinal and lateral directional motion.

The effect of increasing the inclination of Orbital Plane, *i*, can be neglected in the longitudinal motion. In lateral directional motion, increasing the inclination will reduce steady state error of $\varphi_S$ and $\psi_S$.